\documentclass[10pt,twocolumn,letterpaper]{article}

\usepackage{cvpr}              % To produce the CAMERA-READY version
\usepackage{bm}

\newcommand{\rvx}{\mathbf{x}}
\newcommand{\rvy}{\mathbf{y}}
\newcommand{\rvm}{\mathbf{m}}

\newcommand{\methodname}[0]{BlazeEdit\xspace }
\newcommand{\smallsec}[1]{\vspace{0.2em}\noindent\textbf{#1}}

\definecolor{cvprblue}{rgb}{0.21,0.49,0.74}
\usepackage[pagebackref,breaklinks,colorlinks,allcolors=cvprblue]{hyperref}

\def\paperID{2} % *** Enter the Paper ID here
\def\confName{CVPR}
\def\confYear{2026}

\title{BlazeEdit: Generalist Image Editing on Mobile Devices\\with Image-to-Image Diffusion Models}

\author{Fei Deng$^\star$, Yanwu Xu$^\star$, Zhipeng Bao, Zhixing Zhang, Haolin Jia, Karthik Raveendran, Jianing Wei\\[5pt]
Google
}

\begin{document}

\maketitle

\def\thefootnote{$\star$}\footnotetext{Equal contribution.}
\def\thefootnote{\arabic{footnote}} % Resets footnote numbering to digits

\begin{abstract}
The remarkable generation quality of modern diffusion models often comes at the cost of massive parameter counts, which necessitate server-side inference with significant computational costs and potential privacy risks. Consequently, there is growing momentum toward developing efficient on-device alternatives. While recent efforts have optimized text-to-image models for mobile hardware, they remain relatively bulky, typically ranging from 0.5B to 1B parameters. We present \methodname, a highly efficient, generalist image-to-image diffusion model tailored for on-device deployment. By identifying that many practical image editing tasks do not require text-based guidance, we eliminate the text-conditioning components and develop a multi-task architecture that consolidates object removal, outpainting, tone correction, relighting, and sticker generation into a single, compact model of only \textbf{195M} parameters. \methodname achieves a substantial reduction in download size and memory overhead while maintaining competitive generation quality. It completes a full inference pass in just \textbf{290ms} on a Pixel 10, delivering a seamless, privacy-preserving, and lightning-fast experience for generalist image editing on the edge.
\end{abstract}    
\section{Introduction}
\label{sec:intro}

Diffusion and flow-based generative models~\citep{sohl2015deep,ddpm,song2021scorebased,lipman2023flow,albergo2023building,liu2023flow} have achieved remarkable visual quality across a variety of digital content creation domains. However, many state-of-the-art diffusion models~\citep{esser2024scaling} employ heavy multi-modal transformers whose parameter counts can reach up to 20B~\citep{qwen_image}, requiring server-side inference on high-end GPUs/TPUs. This not only incurs significant computational costs, but also raises potential concerns regarding the privacy of uploaded personal photos.

Consequently, there is growing interest in developing on-device diffusion models that can run locally and efficiently. Recent efforts, such as SnapFusion~\citep{snapfusion}, SnapGen~\citep{snapgen}, and MobileDiffusion~\citep{mobilediffusion}, have made impressive strides in reducing the denoiser size and latency for text-to-image generation. Nevertheless, these models remain relatively heavy, as the text encoder increases the total model size. For instance, SnapGen leverages multiple text encoders including CLIP~\citep{clip} and Gemma-2-2B~\citep{gemma2}, adding $\sim$2B parameters to its 0.38B denoiser. The large model size often presents a download barrier for mobile applications, where bandwidth is frequently limited.

In this paper, we challenge the prevailing assumption that a general-purpose mobile editing tool must be built upon a text-to-image foundation. We observe that a significant subset of common image editing actions, such as removing background objects, changing the aspect ratio of an image (\eg, portrait to square or landscape), and creating stylized stickers from photos, can be sufficiently guided by the input image plus a user-provided mask, bypassing the need for text-based conditioning.

Building on this insight, we develop a pure image-to-image framework and introduce \methodname, a highly efficient, generalist image editor tailored for mobile devices. Our contributions are summarized as follows:
\begin{itemize}
    \item We develop a pretraining pipeline specifically for image-to-image diffusion models. By leveraging masked reconstruction as the pretraining objective, we endow the base model with foundational inpainting and outpainting capabilities, facilitating data-efficient downstream finetuning.
    \item We repurpose the mask value as a universal task indicator. Combined with a jointly trained image-and-mask encoder, this allows for simultaneous multi-task finetuning that enables knowledge transfer across tasks.
    \item We achieve a single, compact model of only \textbf{195M} parameters that consolidates five distinct editing tasks---object removal, outpainting, tone correction, relighting, and sticker generation. With an inference latency of just \textbf{290ms} on a Pixel 10, our model delivers a highly interactive user experience.
\end{itemize}
\begin{figure*}[ht]
    \centering
    \includegraphics[width=0.9\textwidth]{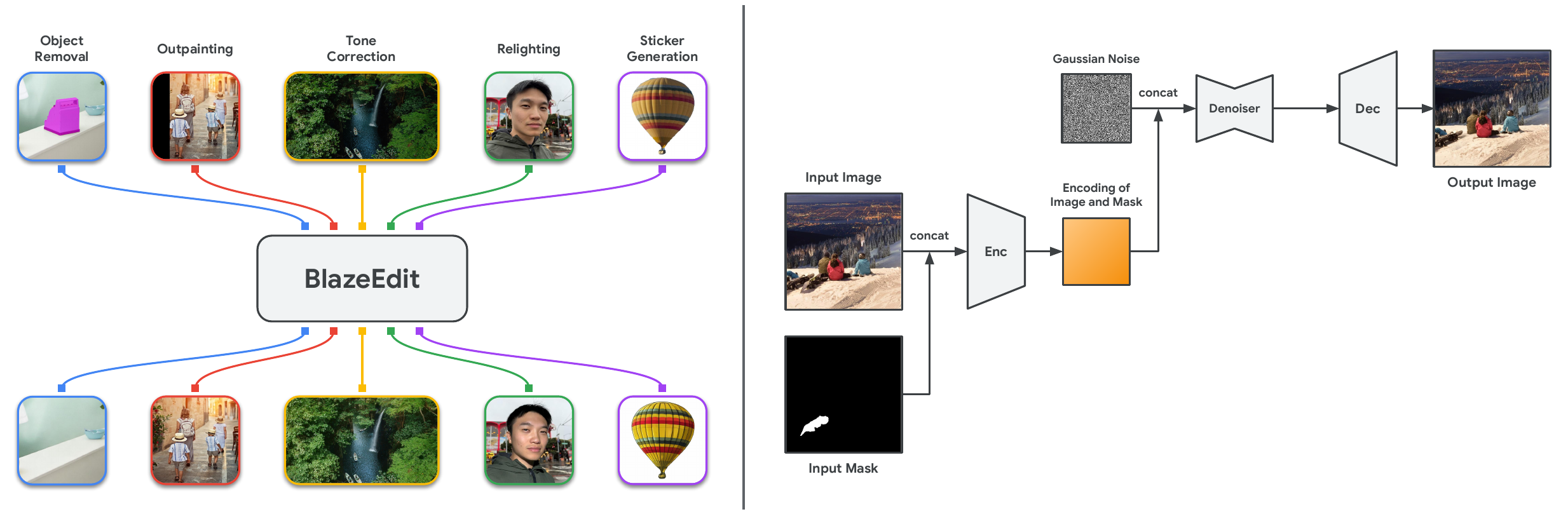}
    \caption{\textit{(Left)} \methodname is a generalist image-to-image editing framework designed for mobile devices. We achieve a single, compact model that consolidates five distinct editing tasks. \textit{(Right)} \methodname follows the latent diffusion paradigm, and employs a trainable image-and-mask encoder to provide the conditioning signal to the denoiser. The input resolution is 512$\times$512, and latent resolution is 64$\times$64.}
    \label{fig:arch}
\end{figure*}

\section{Related Work}
\label{sec:related_work}

There have been significant recent advancements in developing generalist text-to-image diffusion models that unify diverse, multi-modal editing tasks within single architectures~\citep{omnigen,onediffusion,univg,dreamomni,qwen_image,unicontrol,unicontrolnet,versatilediffusion}. However, their reliance on heavy multi-modal transformers and massive parameter counts renders them impractical for on-device execution. A growing body of research has focused on compressing text-to-image diffusion models for mobile devices. Pioneering works such as SnapFusion~\citep{snapfusion}, SnapGen~\citep{snapgen}, and MobileDiffusion~\citep{mobilediffusion} have significantly reduced the model parameters to the 0.5B $\sim$ 1B range through extensive architectural investigation and advanced distillation. Our work further shrinks the model size by shifting to an image-to-image framework that still supports a wide variety of practical editing tasks. Compared to the pixel-space image-to-image framework presented in Palette~\citep{palette}, our model works in latent space and introduces a task-agnostic pretraining phase, enabling fast inference and data-efficient finetuning.
\section{Method}
\label{sec:method}

\subsection{Model Architecture}

Figure~\ref{fig:arch} presents an overview of the \methodname framework. We detail our design below. 

\smallsec{Latent Diffusion with Jointly Trained Encoder.}
Following established practice, we adopt latent diffusion~\citep{ldm} to reduce the computational overhead of the iterative denoising process. Given a frozen encoder $\mathcal{E}$ and decoder $\mathcal{D}$, we seek to train a denoiser $\epsilon_\theta$ that models the conditional distribution $p(\mathcal{E}(\rvy) \mid \rvx, \rvm)$, where $\rvx$, $\rvy$, $\rvm$ denote the input image, output image, and mask, respectively. The prevailing approach in pixel-space image-to-image frameworks, such as Palette~\citep{palette}, provides the masked image as a conditioning input to the denoiser. However, we find this approach suboptimal when adapted to the latent space. Specifically, if the conditioning input is simply changed to the latent representation of the masked image $\mathcal{E}(\rvm \odot \rvx)$, the denoiser will struggle to preserve structural fidelity around the masked region, especially for object removal. This happens even if the autoencoder has no difficulty reconstructing the masked image, \ie, $\mathcal{D}(\mathcal{E}(\rvm \odot \rvx)) \approx \rvm \odot \rvx$. We attribute this to the suboptimal latent representations. Since the autoencoder is optimized mainly for reconstruction, its latent space is not inherently structured to decouple the original image content from the superimposed mask.

To address this, \methodname introduces a secondary, trainable encoder $f_\theta(\texttt{concat}[\rvx, \rvm])$ that processes the concatenated input image and mask. The output of $f_\theta$ is fed to the denoiser as a conditioning input. Unlike the frozen, reconstruction-oriented autoencoder, $f_\theta$ is jointly optimized with the denoiser to extract task-relevant features that the model can more easily leverage. We find this joint training critical for maintaining fine-grained details and structural integrity. Furthermore, for object removal tasks where the mask specifies the object to be removed, providing $f_\theta$ with the full input image $\rvx$ allows the model to more effectively infer and eliminate shadows cast by the object.

\smallsec{Efficient Denoiser and Lightweight Decoder.}
The denoiser follows the U-ViT architecture~\citep{simplediffusion}. Specifically, we employ ResNet~\citep{resnet} blocks at higher resolutions to maintain the spatial inductive bias while saving memory, and self-attention~\citep{transformer} blocks at lower resolutions to capture long-range dependencies and improve accelerator utilization. To further reduce model size and inference latency, we prune the width and depth of the decoder $\mathcal{D}$. We then train it for image reconstruction with the encoder $\mathcal{E}$ frozen, following MobileDiffusion~\citep{mobilediffusion}. The resulting lightweight decoder has only 6M parameters, and shows minimal degradation in reconstruction quality.

\subsection{Pretraining via Masked Reconstruction}

A significant bottleneck in developing a generalist image editor is the scarcity of high-quality, task-specific datasets. Existing literature typically follows one of two paths: (1) adapting a pretrained text-to-image model by adding task-specific LoRAs~\citep{lora,mobilediffusion}, or (2) training an image-to-image model from scratch but limited to a small number of tasks, such as colorization and JPEG restoration~\citep{palette}, where massive paired data can be easily synthesized. The first approach inevitably involves parameters for text understanding and text-image alignment, which are unnecessary for purely image-conditioned tasks. On the other hand, the second approach struggles to scale to more complex editing tasks where the training data are costly to obtain.

To achieve data-efficient scaling across diverse editing tasks while maintaining a compact model footprint, we propose to first pretrain an image-to-image base model from scratch on large-scale task-agnostic datasets, and then finetune it jointly on all available task-specific datasets.

Inspired by the masked image modeling framework~\citep{mae,maskgit}, we use masked reconstruction as our pretraining objective. We find it critical to employ a diverse set of masks, including random patches, geometric shapes, and strokes within the image, and paddings on the boundary of the image. This not only encourages the model to learn an expressive image representation, but also equips the model with core inpainting and outpainting capabilities, building a generalizable foundation for specialized downstream tasks.

We directly leverage the images from existing large-scale text-to-image datasets for our pretraining. The loss function can be written as:
\begin{align}
    \mathbb{E}_{\rvx, \rvm, \bm{\epsilon}, t} \lVert \epsilon_\theta(\tilde{\rvx}_t, f_\theta(\texttt{concat}[\rvm \odot \rvx, \rvm]), t) - \bm{\epsilon} \rVert^2\ ,
\end{align}
where $\bm{\epsilon} \sim \mathcal{N}(\bm{0}, \bm{1})$ denotes the random noise, and $\tilde{\rvx}_t$ is the noised version of $\mathcal{E}(\rvx)$ at a randomly sampled diffusion step $t$. The mask $\rvm$ is randomly generated on the fly. To prevent degenerate solutions caused by information leak, we mask the image $\rvx$ when feeding it to the trainable encoder $f_\theta$. This is only necessary during pretraining.

\subsection{Multi-Task Finetuning and Distillation}

Following pretraining on large-scale, general-purpose image data, we transition to supervised finetuning and step distillation on task-specific datasets.

\smallsec{Universal Task Signaling.}
We conduct multi-task finetuning on all available downstream tasks simultaneously, as this enables knowledge transfer across tasks. To help the model distinguish between different tasks, we introduce a universal task signaling mechanism encoded directly within the mask values. Specifically, each task $i$ is assigned a unique numerical constant $\tau_i$, which is used to scale the binary mask $\rvm$. This mask-based conditioning signal allows the model to switch functional modes dynamically without additional parameters. For an input-mask-output triplet $(\rvx, \rvm, \rvy)$ belonging to task $i$, the finetuning objective is formulated as:
\begin{align}
    \mathbb{E}_{\bm{\epsilon}, t} \lVert \epsilon_\theta(\tilde{\rvy}_t, f_\theta(\texttt{concat}[\rvx, \tau_i \cdot \rvm]), t) - \bm{\epsilon} \rVert^2\ ,
\end{align}
where $\bm{\epsilon} \sim \mathcal{N}(\bm{0}, \bm{1})$ denotes the random noise, and $\tilde{\rvy}_t$ is the noised version of $\mathcal{E}(\rvy)$ at a randomly sampled diffusion step $t$.

\smallsec{Adversarial Distribution Matching Distillation.}
After supervised finetuning, we distill the resulting model for 2-step inference, significantly reducing latency. We find that distribution matching distillation~\citep{dmd} combined with adversarial training~\citep{dmd2} works well in this few-step regime.

\section{Experiments}
\label{sec:experiment}

\begin{figure*}[ht]
    \centering
    \includegraphics[width=0.9\textwidth]{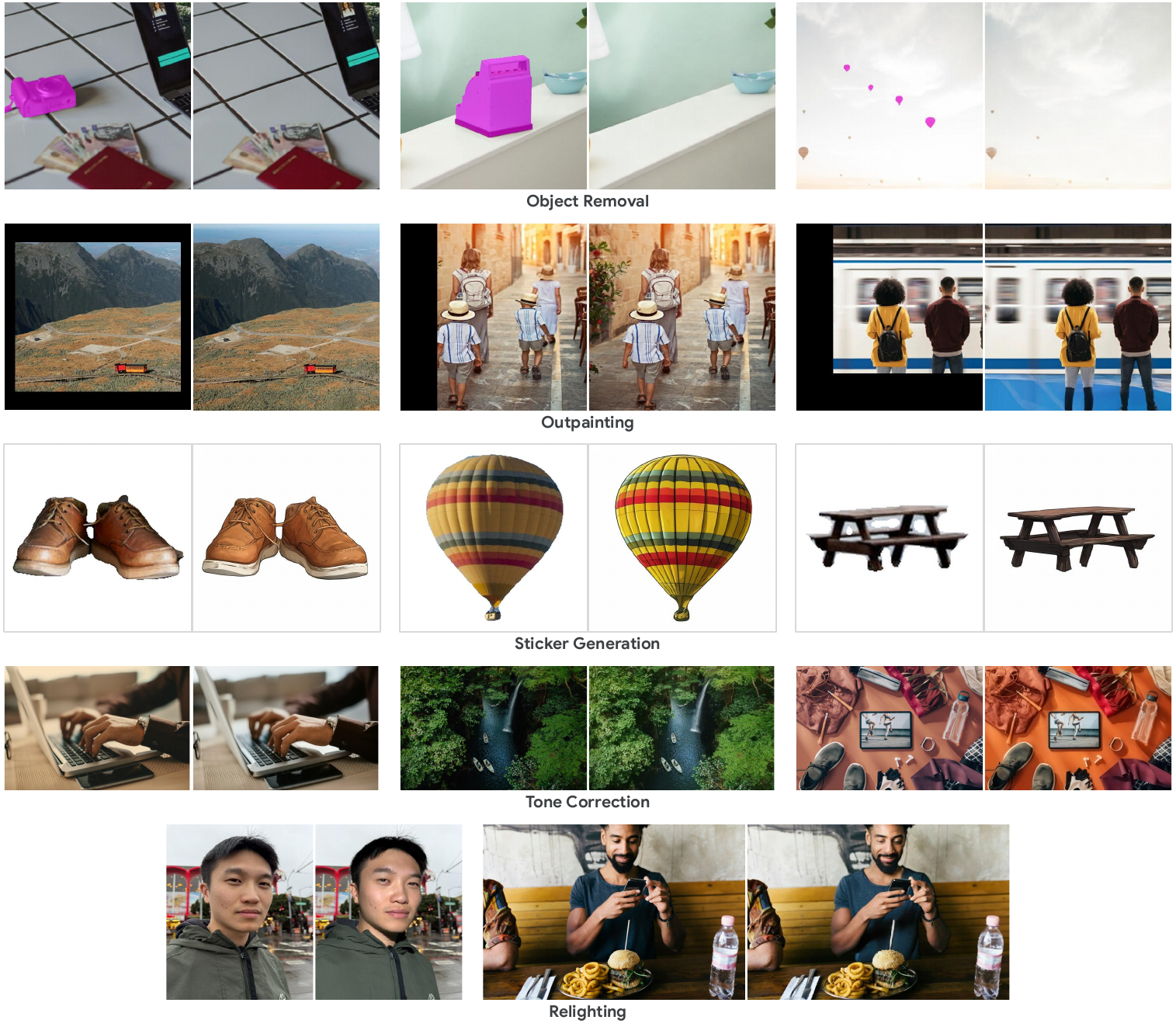}
    \vspace{-5pt}
    \caption{Qualitative results of \methodname on five distinct editing tasks. \methodname achieves compelling and versatile image-to-image editing. It exhibits strong structural reasoning and semantic preservation, seamlessly handling object and shadow removal, boundary extrapolation, stylization, tone correction, and lighting adjustment.}
    \label{fig:result}
\end{figure*}

\subsection{Tasks and Datasets}

We first describe the range of image editing tasks supported by \methodname and the task-specific finetuning datasets.

\smallsec{Object Removal}
aims to remove a user-specified object from a scene and synthesize a plausible background in its place, helping to reduce visual clutter. We use a manually curated dataset of $\sim$20K image pairs. Following \citep{objectdrop}, each pair captures a scene before and after an object is physically removed while minimizing other changes.

\smallsec{Outpainting}
extends the boundaries of an image, which is useful for changing its aspect ratio (\eg, portrait to landscape). We use a high-quality subset of $\sim$5K images from our pretraining datasets, filtered for aesthetic and diversity.

\smallsec{Tone Correction}
enhances an image by refining its white balance, saturation, and exposure. We synthesize a dataset of $\sim$3M image pairs by a high-performance teacher model.

\smallsec{Relighting}
focuses on portrait enhancement by mitigating unfavorable lighting, such as harsh facial shadows. We use a dataset of $\sim$100K image pairs synthesized by applying diverse shadow augmentations to portrait photography~\citep{futschik2023controllable}.

\smallsec{Sticker Generation}
stylizes user-specified subjects within an image (\eg, people, pets, and objects), transforming them into high-quality artistic stickers. We synthesize a diverse dataset of $\sim$100K image pairs using a high-capacity text-to-image generation model adept at identity-preserving stylization.

\subsection{Main Results}

\smallsec{Qualitative Results.}
In Figure~\ref{fig:result}, we present the input images and generation samples from \methodname on all five tasks. \methodname achieves compelling and versatile image-to-image editing. It exhibits strong structural reasoning and semantic preservation, seamlessly handling object and shadow removal, boundary extrapolation, stylization, tone correction, and lighting adjustment.

\smallsec{Efficiency Evaluation.}
In Table~\ref{tab:size}, we compare the model footprint of \methodname with prior works. \methodname achieves 2$\times$ reduction in denoiser parameter count while eliminating the text encoders, which typically range from 0.1B to 2B parameters. Furthermore, Table~\ref{tab:latency} provides a breakdown of inference latency on Pixel 10 using Edge TPU. \methodname completes a full inference pass in just 290ms, enabling lightning-fast and privacy-preserving image editing directly on the edge.

\begin{table}[t]
    \centering
    \caption{Model footprint comparison. \methodname achieves 2$\times$ reduction in denoiser parameters while eliminating text encoders. }
    \vspace{-0.1in}
    \resizebox{\linewidth}{!}{
    \begin{tabular}{l|ccccc}
    \toprule
        Model & Text Encoder & Denoiser \#Params \\
    \midrule
        SnapFusion~\citep{snapfusion} & CLIP-ViT-H & 848M \\
        MobileDiffusion~\citep{mobilediffusion} & CLIP-ViT-L & 386M \\
        SnapGen~\citep{snapgen} & CLIP-ViT-L, CLIP-ViT-G, Gemma-2-2B & 379M \\
    \midrule
        \methodname & None & 189M \\
    \bottomrule 
    \end{tabular}
    }
    \label{tab:size}
\end{table}

\begin{table}[t]
    \centering
    \caption{Inference latency measurements. \methodname completes a full inference pass in just 290ms, enabling lightning-fast and privacy-preserving image editing directly on the edge.}
    \vspace{-0.1in}
    \resizebox{\linewidth}{!}{
    \begin{tabular}{c|ccc|c}
    \toprule 
       Device  & Encoder & Decoder & Denoiser (2 steps) & Overall \\
    \midrule
       Pixel 10 (Edge TPU) & 45ms & 55ms & 190ms & 290ms \\
    \bottomrule
    \end{tabular}
    }
    \label{tab:latency}
    \vspace{-0.16in}
\end{table}
\section{Conclusion}
\label{sec:conclusion}

We presented \methodname, a highly efficient image-to-image diffusion model designed for edge-native generalist image editing. We successfully consolidated five diverse editing tasks into a unified 195M-parameter architecture. Our experiments demonstrate that \methodname achieves competitive results while delivering highly interactive user experience on mobile hardware.

\section*{Acknowledgements}
We thank Julius Kammerl, Daniel Fenner, Lutz Justen, Florian Kübler, Ronald Wotzlaw, Steven Toribio, Weiyi Wang, Marissa Ikonomidis, Lu Wang, Juhyun Lee, Zahra Koochak, Gregory Karpiak, Renjie Wu, Joe Zou, Pauline Sho, and Advait Jain for their essential support with on-device pipeline implementation, model export, quantization, and performance optimization. We also appreciate Yang Zhao, Zhisheng Xiao, Oliver Wang, Dana Berman, Dani Lischinski, Ron Shapiro, Tomer Golany, Alex Rav Acha, Michael Milne, Gilles Roux, Zarana Parekh, James Vecore, Jason Chang, Mogan Shieh, and Chris Parsons for their valuable insights and data support.

{
    \small
    \bibliographystyle{ieeenat_fullname}
    \bibliography{main}

@inproceedings{snapfusion,
  title={{SnapFusion}: Text-to-image diffusion model on mobile devices within two seconds},
  author={Li, Yanyu and Wang, Huan and Jin, Qing and Hu, Ju and Chemerys, Pavlo and Fu, Yun and Wang, Yanzhi and Tulyakov, Sergey and Ren, Jian},
  booktitle={Advances in Neural Information Processing Systems},
  year={2023}
}

@inproceedings{mobilediffusion,
  title={{MobileDiffusion}: Instant text-to-image generation on mobile devices},
  author={Zhao, Yang and Xu, Yanwu and Xiao, Zhisheng and Jia, Haolin and Hou, Tingbo},
  booktitle={European Conference on Computer Vision},
  pages={225--242},
  year={2024},
  organization={Springer}
}

@inproceedings{palette,
  title={Palette: Image-to-image diffusion models},
  author={Saharia, Chitwan and Chan, William and Chang, Huiwen and Lee, Chris and Ho, Jonathan and Salimans, Tim and Fleet, David and Norouzi, Mohammad},
  booktitle={ACM SIGGRAPH 2022 Conference Proceedings},
  pages={1--10},
  year={2022}
}

@inproceedings{objectdrop,
  title={{ObjectDrop}: Bootstrapping counterfactuals for photorealistic object removal and insertion},
  author={Winter, Daniel and Cohen, Matan and Fruchter, Shlomi and Pritch, Yael and Rav-Acha, Alex and Hoshen, Yedid},
  booktitle={European Conference on Computer Vision},
  pages={112--129},
  year={2024},
  organization={Springer}
}

@inproceedings{simplediffusion,
  title={{simple diffusion}: End-to-end diffusion for high resolution images},
  author={Hoogeboom, Emiel and Heek, Jonathan and Salimans, Tim},
  booktitle={International Conference on Machine Learning},
  year={2023}
}

@InProceedings{sohl2015deep,
  title = 	 {Deep Unsupervised Learning using Nonequilibrium Thermodynamics},
  author = 	 {Sohl-Dickstein, Jascha and Weiss, Eric and Maheswaranathan, Niru and Ganguli, Surya},
  booktitle = 	 {International Conference on Machine Learning},
  year = 	 {2015}
}

@inproceedings{ddpm,
 author = {Ho, Jonathan and Jain, Ajay and Abbeel, Pieter},
 booktitle = {Advances in Neural Information Processing Systems},
 title = {Denoising Diffusion Probabilistic Models},
 year = {2020}
}

@inproceedings{song2021scorebased,
title={Score-Based Generative Modeling through Stochastic Differential Equations},
author={Yang Song and Jascha Sohl-Dickstein and Diederik P Kingma and Abhishek Kumar and Stefano Ermon and Ben Poole},
booktitle={International Conference on Learning Representations},
year={2021}
}

@inproceedings{lipman2023flow,
title={Flow Matching for Generative Modeling},
author={Yaron Lipman and Ricky T. Q. Chen and Heli Ben-Hamu and Maximilian Nickel and Matthew Le},
booktitle={International Conference on Learning Representations},
year={2023},
url={https://openreview.net/forum?id=PqvMRDCJT9t}
}

@inproceedings{albergo2023building,
title={Building Normalizing Flows with Stochastic Interpolants},
author={Michael Samuel Albergo and Eric Vanden-Eijnden},
booktitle={International Conference on Learning Representations},
year={2023},
url={https://openreview.net/forum?id=li7qeBbCR1t}
}

@inproceedings{liu2023flow,
title={Flow Straight and Fast: Learning to Generate and Transfer Data with Rectified Flow},
author={Xingchao Liu and Chengyue Gong and Qiang Liu},
booktitle={International Conference on Learning Representations},
year={2023},
url={https://openreview.net/forum?id=XVjTT1nw5z}
}

@inproceedings{esser2024scaling,
title={Scaling Rectified Flow Transformers for High-Resolution Image Synthesis},
author={Patrick Esser and Sumith Kulal and Andreas Blattmann and Rahim Entezari and Jonas M{\"u}ller and Harry Saini and Yam Levi and Dominik Lorenz and Axel Sauer and Frederic Boesel and Dustin Podell and Tim Dockhorn and Zion English and Robin Rombach},
booktitle={International Conference on Machine Learning},
year={2024},
url={https://openreview.net/forum?id=FPnUhsQJ5B}
}

@misc{qwen_image,
      title={{Qwen-Image} Technical Report}, 
      author={Chenfei Wu and Jiahao Li and Jingren Zhou and Junyang Lin and Kaiyuan Gao and Kun Yan and Sheng-ming Yin and Shuai Bai and Xiao Xu and Yilei Chen and Yuxiang Chen and Zecheng Tang and Zekai Zhang and Zhengyi Wang and An Yang and Bowen Yu and Chen Cheng and Dayiheng Liu and Deqing Li and Hang Zhang and Hao Meng and Hu Wei and Jingyuan Ni and Kai Chen and Kuan Cao and Liang Peng and Lin Qu and Minggang Wu and Peng Wang and Shuting Yu and Tingkun Wen and Wensen Feng and Xiaoxiao Xu and Yi Wang and Yichang Zhang and Yongqiang Zhu and Yujia Wu and Yuxuan Cai and Zenan Liu},
      year={2025},
      eprint={2508.02324},
      archivePrefix={arXiv},
      primaryClass={cs.CV},
      url={https://arxiv.org/abs/2508.02324}, 
}

@inproceedings{mae,
  title={Masked autoencoders are scalable vision learners},
  author={He, Kaiming and Chen, Xinlei and Xie, Saining and Li, Yanghao and Doll{\'a}r, Piotr and Girshick, Ross},
  booktitle={Proceedings of the IEEE/CVF Conference on Computer Vision and Pattern Recognition},
  pages={16000--16009},
  year={2022}
}

@inproceedings{maskgit,
  title={{MaskGIT}: Masked generative image transformer},
  author={Chang, Huiwen and Zhang, Han and Jiang, Lu and Liu, Ce and Freeman, William T},
  booktitle={Proceedings of the IEEE/CVF Conference on Computer Vision and Pattern Recognition},
  pages={11315--11325},
  year={2022}
}

@inproceedings{ldm,
  title={High-resolution image synthesis with latent diffusion models},
  author={Rombach, Robin and Blattmann, Andreas and Lorenz, Dominik and Esser, Patrick and Ommer, Bj{\"o}rn},
  booktitle={Proceedings of the IEEE/CVF Conference on Computer Vision and Pattern Recognition},
  pages={10684--10695},
  year={2022}
}

@inproceedings{resnet,
  title={Deep residual learning for image recognition},
  author={He, Kaiming and Zhang, Xiangyu and Ren, Shaoqing and Sun, Jian},
  booktitle={Proceedings of the IEEE Conference on Computer Vision and Pattern Recognition},
  pages={770--778},
  year={2016}
}

@inproceedings{transformer,
  title={Attention is all you need},
  author={Vaswani, Ashish and Shazeer, Noam and Parmar, Niki and Uszkoreit, Jakob and Jones, Llion and Gomez, Aidan N and Kaiser, {\L}ukasz and Polosukhin, Illia},
  booktitle={Advances in neural information processing systems},
  year={2017}
}

@InProceedings{snapgen,
    author    = {Chen, Jierun and Hu, Dongting and Huang, Xijie and Coskun, Huseyin and Sahni, Arpit and Gupta, Aarush and Goyal, Anujraaj and Lahiri, Dishani and Singh, Rajesh and Idelbayev, Yerlan and Cao, Junli and Li, Yanyu and Cheng, Kwang-Ting and Chan, S.-H. Gary and Gong, Mingming and Tulyakov, Sergey and Kag, Anil and Xu, Yanwu and Ren, Jian},
    title     = {{SnapGen}: Taming High-Resolution Text-to-Image Models for Mobile Devices with Efficient Architectures and Training},
    booktitle = {Proceedings of the IEEE/CVF Conference on Computer Vision and Pattern Recognition},
    year      = {2025},
    pages     = {7997-8008}
}

@inproceedings{lora,
title={Lo{RA}: Low-Rank Adaptation of Large Language Models},
author={Edward J Hu and yelong shen and Phillip Wallis and Zeyuan Allen-Zhu and Yuanzhi Li and Shean Wang and Lu Wang and Weizhu Chen},
booktitle={International Conference on Learning Representations},
year={2022},
url={https://openreview.net/forum?id=nZeVKeeFYf9}
}

@inproceedings{dmd,
  title={One-step diffusion with distribution matching distillation},
  author={Yin, Tianwei and Gharbi, Micha{\"e}l and Zhang, Richard and Shechtman, Eli and Durand, Fredo and Freeman, William T and Park, Taesung},
  booktitle={Proceedings of the IEEE/CVF Conference on Computer Vision and Pattern Recognition},
  pages={6613--6623},
  year={2024}
}

@inproceedings{dmd2,
  title={Improved distribution matching distillation for fast image synthesis},
  author={Yin, Tianwei and Gharbi, Micha{\"e}l and Park, Taesung and Zhang, Richard and Shechtman, Eli and Durand, Fredo and Freeman, Bill},
  booktitle={Advances in neural information processing systems},
  year={2024}
}

@misc{gemma2,
      title={Gemma 2: Improving Open Language Models at a Practical Size}, 
      author={Gemma Team and Morgane Riviere and Shreya Pathak and Pier Giuseppe Sessa and Cassidy Hardin and Surya Bhupatiraju and Léonard Hussenot and Thomas Mesnard and Bobak Shahriari and Alexandre Ramé and Johan Ferret and Peter Liu and Pouya Tafti and Abe Friesen and Michelle Casbon and Sabela Ramos and Ravin Kumar and Charline Le Lan and Sammy Jerome and Anton Tsitsulin and Nino Vieillard and Piotr Stanczyk and Sertan Girgin and Nikola Momchev and Matt Hoffman and Shantanu Thakoor and Jean-Bastien Grill and Behnam Neyshabur and Olivier Bachem and Alanna Walton and Aliaksei Severyn and Alicia Parrish and Aliya Ahmad and Allen Hutchison and Alvin Abdagic and Amanda Carl and Amy Shen and Andy Brock and Andy Coenen and Anthony Laforge and Antonia Paterson and Ben Bastian and Bilal Piot and Bo Wu and Brandon Royal and Charlie Chen and Chintu Kumar and Chris Perry and Chris Welty and Christopher A. Choquette-Choo and Danila Sinopalnikov and David Weinberger and Dimple Vijaykumar and Dominika Rogozińska and Dustin Herbison and Elisa Bandy and Emma Wang and Eric Noland and Erica Moreira and Evan Senter and Evgenii Eltyshev and Francesco Visin and Gabriel Rasskin and Gary Wei and Glenn Cameron and Gus Martins and Hadi Hashemi and Hanna Klimczak-Plucińska and Harleen Batra and Harsh Dhand and Ivan Nardini and Jacinda Mein and Jack Zhou and James Svensson and Jeff Stanway and Jetha Chan and Jin Peng Zhou and Joana Carrasqueira and Joana Iljazi and Jocelyn Becker and Joe Fernandez and Joost van Amersfoort and Josh Gordon and Josh Lipschultz and Josh Newlan and Ju-yeong Ji and Kareem Mohamed and Kartikeya Badola and Kat Black and Katie Millican and Keelin McDonell and Kelvin Nguyen and Kiranbir Sodhia and Kish Greene and Lars Lowe Sjoesund and Lauren Usui and Laurent Sifre and Lena Heuermann and Leticia Lago and Lilly McNealus and Livio Baldini Soares and Logan Kilpatrick and Lucas Dixon and Luciano Martins and Machel Reid and Manvinder Singh and Mark Iverson and Martin Görner and Mat Velloso and Mateo Wirth and Matt Davidow and Matt Miller and Matthew Rahtz and Matthew Watson and Meg Risdal and Mehran Kazemi and Michael Moynihan and Ming Zhang and Minsuk Kahng and Minwoo Park and Mofi Rahman and Mohit Khatwani and Natalie Dao and Nenshad Bardoliwalla and Nesh Devanathan and Neta Dumai and Nilay Chauhan and Oscar Wahltinez and Pankil Botarda and Parker Barnes and Paul Barham and Paul Michel and Pengchong Jin and Petko Georgiev and Phil Culliton and Pradeep Kuppala and Ramona Comanescu and Ramona Merhej and Reena Jana and Reza Ardeshir Rokni and Rishabh Agarwal and Ryan Mullins and Samaneh Saadat and Sara Mc Carthy and Sarah Cogan and Sarah Perrin and Sébastien M. R. Arnold and Sebastian Krause and Shengyang Dai and Shruti Garg and Shruti Sheth and Sue Ronstrom and Susan Chan and Timothy Jordan and Ting Yu and Tom Eccles and Tom Hennigan and Tomas Kocisky and Tulsee Doshi and Vihan Jain and Vikas Yadav and Vilobh Meshram and Vishal Dharmadhikari and Warren Barkley and Wei Wei and Wenming Ye and Woohyun Han and Woosuk Kwon and Xiang Xu and Zhe Shen and Zhitao Gong and Zichuan Wei and Victor Cotruta and Phoebe Kirk and Anand Rao and Minh Giang and Ludovic Peran and Tris Warkentin and Eli Collins and Joelle Barral and Zoubin Ghahramani and Raia Hadsell and D. Sculley and Jeanine Banks and Anca Dragan and Slav Petrov and Oriol Vinyals and Jeff Dean and Demis Hassabis and Koray Kavukcuoglu and Clement Farabet and Elena Buchatskaya and Sebastian Borgeaud and Noah Fiedel and Armand Joulin and Kathleen Kenealy and Robert Dadashi and Alek Andreev},
      year={2024},
      eprint={2408.00118},
      archivePrefix={arXiv},
      primaryClass={cs.CL},
      url={https://arxiv.org/abs/2408.00118}, 
}

@InProceedings{clip,
  title = 	 {Learning Transferable Visual Models From Natural Language Supervision},
  author =       {Radford, Alec and Kim, Jong Wook and Hallacy, Chris and Ramesh, Aditya and Goh, Gabriel and Agarwal, Sandhini and Sastry, Girish and Askell, Amanda and Mishkin, Pamela and Clark, Jack and Krueger, Gretchen and Sutskever, Ilya},
  booktitle = 	 {International Conference on Machine Learning},
  year = 	 {2021}
}

@inproceedings{omnigen,
  title={{OmniGen}: Unified image generation},
  author={Xiao, Shitao and Wang, Yueze and Zhou, Junjie and Yuan, Huaying and Xing, Xingrun and Yan, Ruiran and Li, Chaofan and Wang, Shuting and Huang, Tiejun and Liu, Zheng},
  booktitle={Proceedings of the IEEE/CVF Conference on Computer Vision and Pattern Recognition},
  pages={13294--13304},
  year={2025}
}

@InProceedings{onediffusion,
    author    = {Le, Duong H. and Pham, Tuan and Lee, Sangho and Clark, Christopher and Kembhavi, Aniruddha and Mandt, Stephan and Krishna, Ranjay and Lu, Jiasen},
    title     = {One Diffusion to Generate Them All},
    booktitle = {Proceedings of the IEEE/CVF Conference on Computer Vision and Pattern Recognition},
    year      = {2025},
    pages     = {2671-2682}
}

@InProceedings{univg,
    author    = {Fu, Tsu-Jui and Qian, Yusu and Chen, Chen and Hu, Wenze and Gan, Zhe and Yang, Yinfei},
    title     = {{UniVG}: A Generalist Diffusion Model for Unified Image Generation and Editing},
    booktitle = {Proceedings of the IEEE/CVF International Conference on Computer Vision},
    year      = {2025},
    pages     = {17160-17170}
}

@inproceedings{unicontrol,
 author = {Qin, Can and Zhang, Shu and Yu, Ning and Feng, Yihao and Yang, Xinyi and Zhou, Yingbo and Wang, Huan and Niebles, Juan Carlos and Xiong, Caiming and Savarese, Silvio and Ermon, Stefano and Fu, Yun and Xu, Ran},
 booktitle = {Advances in Neural Information Processing Systems},
 title = {{UniControl}: A Unified Diffusion Model for Controllable Visual Generation In the Wild},
 year = {2023}
}

@InProceedings{versatilediffusion,
    author    = {Xu, Xingqian and Wang, Zhangyang and Zhang, Gong and Wang, Kai and Shi, Humphrey},
    title     = {{Versatile Diffusion}: Text, Images and Variations All in One Diffusion Model},
    booktitle = {Proceedings of the IEEE/CVF International Conference on Computer Vision},
    year      = {2023},
    pages     = {7754-7765}
}

@inproceedings{unicontrolnet,
  title={{Uni-ControlNet}: All-in-one control to text-to-image diffusion models},
  author={Zhao, Shihao and Chen, Dongdong and Chen, Yen-Chun and Bao, Jianmin and Hao, Shaozhe and Yuan, Lu and Wong, Kwan-Yee K},
  booktitle={Advances in neural information processing systems},
  year={2023}
}

@InProceedings{dreamomni,
    author    = {Xia, Bin and Zhang, Yuechen and Li, Jingyao and Wang, Chengyao and Wang, Yitong and Wu, Xinglong and Yu, Bei and Jia, Jiaya},
    title     = {{DreamOmni}: Unified Image Generation and Editing},
    booktitle = {Proceedings of the IEEE/CVF Conference on Computer Vision and Pattern Recognition},
    year      = {2025},
    pages     = {28533-28543}
}

@inproceedings{futschik2023controllable,
  title={Controllable light diffusion for portraits},
  author={Futschik, David and Ritland, Kelvin and Vecore, James and Fanello, Sean and Orts-Escolano, Sergio and Curless, Brian and S{\`y}kora, Daniel and Pandey, Rohit},
  booktitle={Proceedings of the IEEE/CVF Conference on Computer Vision and Pattern Recognition},
  pages={8412--8421},
  year={2023}
}
}

% WARNING: do not forget to delete the supplementary pages from your submission 
% \input{sec/X_suppl}

\end{document}